\documentclass[conference]{IEEEtran}
\IEEEoverridecommandlockouts
\usepackage{cite}
\usepackage{amsmath,amssymb,amsfonts}
\usepackage{algorithmic}
\usepackage{graphicx}
\usepackage{comment}
\usepackage{textcomp}
\usepackage{xcolor}
\usepackage{color}
\usepackage{lipsum}
\usepackage{times}
\usepackage{textcomp}
\usepackage{amsmath,amssymb,amsfonts}
\usepackage[inline]{enumitem}
\usepackage[ruled,vlined,linesnumbered]{algorithm2e}
\usepackage{color}
\usepackage{tablefootnote}
\usepackage{mathtools}

\usepackage{amsthm}
\usepackage{imakeidx}
\usepackage{float}
\usepackage{multicol}
\usepackage{booktabs} 
\usepackage{listings}
\usepackage{verbatim}
\usepackage{amsfonts}
\usepackage{hyperref}
\usepackage{breakurl}
\usepackage{url}

\newcommand{\overbar}[1]{\mkern 1.5mu\overline{\mkern-1.5mu#1\mkern-1.5mu}\mkern 1.5mu}

\makeatletter

\def\BibTeX{{\rm B\kern-.05em{\sc i\kern-.025em b}\kern-.08em
    T\kern-.1667em\lower.7ex\hbox{E}\kern-.125emX}}
\begin{document}

\title{ Leveraging Decentralized Artificial Intelligence to Enhance  Resilience of Energy Networks 
}

\author{\IEEEauthorblockN{Ahmed Imteaj$^{1}$, M. Hadi Amini$^1$, and  Javad Mohammadi$^2$}
\IEEEauthorblockA{{$^1$ School of Computing and Information Sciences, College of Engineering and Computing, Florida International University} \\
{Sustainability, Optimization, and Learning for InterDependent networks laboratory (solid lab), Miami, FL, USA}\\\textit{$^2$} Department of Electrical and Computer Engineering, Carnegie Mellon University, Pittsburgh, PA, USA \\email: aimte001@fiu.edu, moamini@fiu.edu, jmohamma@andrew.cmu.edu}}

\maketitle

\begin{abstract}
This paper reintroduces the notion of resilience in the context of recent issues originated from climate change triggered events including severe hurricanes and wildfires. A recent example is PG\&E's forced power outage to contain wildfire risk which led to widespread power disruption. This paper focuses on answering two questions: \textit{who is responsible for resilience?} and \textit{how to quantify the monetary value of resilience?} To this end, we first provide preliminary definitions of resilience for power systems. We then investigate the role of natural hazards, especially wildfire, on power system resilience. Finally, we will propose a decentralized strategy for a resilient management system using distributed storage and demand response resources. Our proposed high fidelity model provides utilities, operators, and policymakers with a clearer picture for strategic decision making and  preventive decisions.
\end{abstract}


\section{Introduction}
\subsection{Motivation}
Modern societies are heavily relying on continuous power supply provided by the electric grids. Climate change, natural disasters, and cyber threats are challenging the reliability and resilience of the aging power network. 
All infrastructure networks, especially urban areas are highly dependent on the electric grid power-supply \cite{amini2019distributed,amini2019sustainable} and hence, the outage of such a power system can cause an enormous cost which can be considered as an utmost global concern. Besides, extreme natural disasters like wildfire, hurricanes, storms, and heatwaves are just a part of the nation's DNA and these cause power outages impacting the normal operations of companies.
\begin{figure*}[t]
\setlength{\belowcaptionskip}{-20pt}
\begin{center}
  \includegraphics[width=10.5cm]{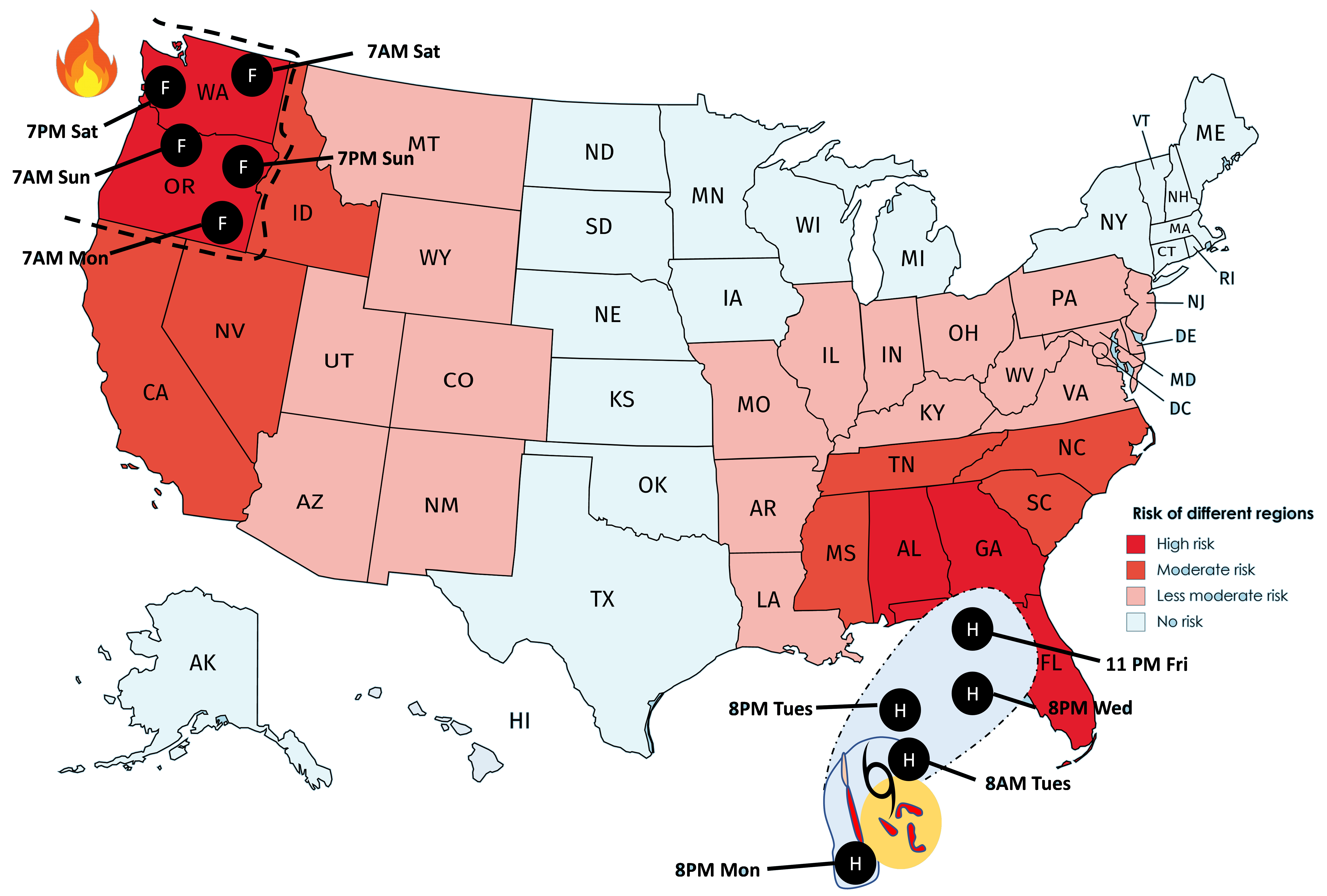}
   \caption{Visualization of fire and hurricane occurrence event on map (H: Hurricane F: Wildfire)}
    \label{fig:1}  
    \end{center}
\end{figure*}
A recent report by The Wall Street Journal claimed that one of the largest utilities of nations named PG\&E is on the edge of bankruptcy because the company's components are responsible for more than 1500 fires in recent years \cite{PG&E}. Breaking out of fire one day on average, causing dozens of deaths, and responsible for the destruction of thousands of acres, the company now has the liability of more than \$30 billion according to a recent analysis. A recent survey found that power outage occurs in one in four companies at least once a month \cite{PGE}. Particularly, for large companies, power interruption loss exceeds a million dollars per hour and the annual forfeiture is around \$150 billion \cite{Bloom}. Hence, our main motivation is to mitigate such costs with the power of intelligence. We provide an outline of an adaptive resilience-based approach where the system will be resistant to power disturbance and able to cluster the risky zone from the main station. Resilience is a multidimensional attribute of a smart grid that means handling disturbances in power supply caused by natural hazards, cyber-physical incidents, or human interventions/attacks. The preeminence of resilience value in the power system with the perspective of natural disasters is one of the main goals of this paper.  
\subsection{Literature Review}
In \cite{amini2013load}, multi-agent-based load management was discussed that counted generation, grid, and demand response to mitigate the peak load. But, their proposed method was based on a centralized approach and agents could not interact among themselves to share resources. The author in \cite{bahrami2018decentralized} proposed a technique to interact among distributed network operator (DNO), load aggregator and generator through which load will be aggregated and power will be distributed based on the energy demand. A decentralized demand response framework was designed in \cite{bahrami2017decentralized}, where the operator handles the interaction between a consumer and supplier, but they did not discuss system outage or resilience mechanism. The fundamental viewpoint and various scope of resilience practical implementations are discussed in \cite{gholami2018toward}. They highlighted the resilience enhancement strategy and showed the relation between resilience privacy, reliability, and security. A detailed survey about multi-agent systems was analyzed in \cite{sujil2018multi} that focused on distributed intelligence of agents that can communicate with each other by participating and managing their demand and load to achieve their target. Besides, a study of a multi-agent based decision support system based on recent applications to the related fields was presented in \cite{salgueiro2020multi}, where they discussed how intelligence can be applied through sensing environment in making decisions from a different point of views. 
A multi-agent based decentralized autonomous smart grid with communication constraints was investigated in \cite{liu2014decentralized}, where the agent can establish communication with neighboring agents considering system frequency. 
\subsection{Contribution}
To minimize the loss that precipitates due to power outages,  smart grids need to be resilient, reliable and robust. In this paper, we explore a visionary solution to address the quantification of resilience value from the perspective of natural menace, discuss possible causes of such outages, and propose a fidelity model that can assist in making prudent decisions during the inconvenience of power systems. To this end, we propose a tri-layer solution. It starts with a classic definition of resilience. In the first layer, we revisit this notion by taking into account artificial intelligence (AI) as a promising means of enabling local energy supply in various stages of the system: pre-event, during the event, and post-event scenarios. In the second layer, we discuss the tools that can be used to implement a decentralized intelligent system that can sense the environment and take necessary actions to mitigate the loss. Finally, the third layer explores the technique to disconnect a certain area that has a high chance to be affected and supply partial electricity without using the direct power supply from the main grid. This layer leverages distributed storage and demand response.

\begin{figure*}[t]
\setlength{\belowcaptionskip}{-20pt}
\begin{center}
  \includegraphics[width=0.74\linewidth]{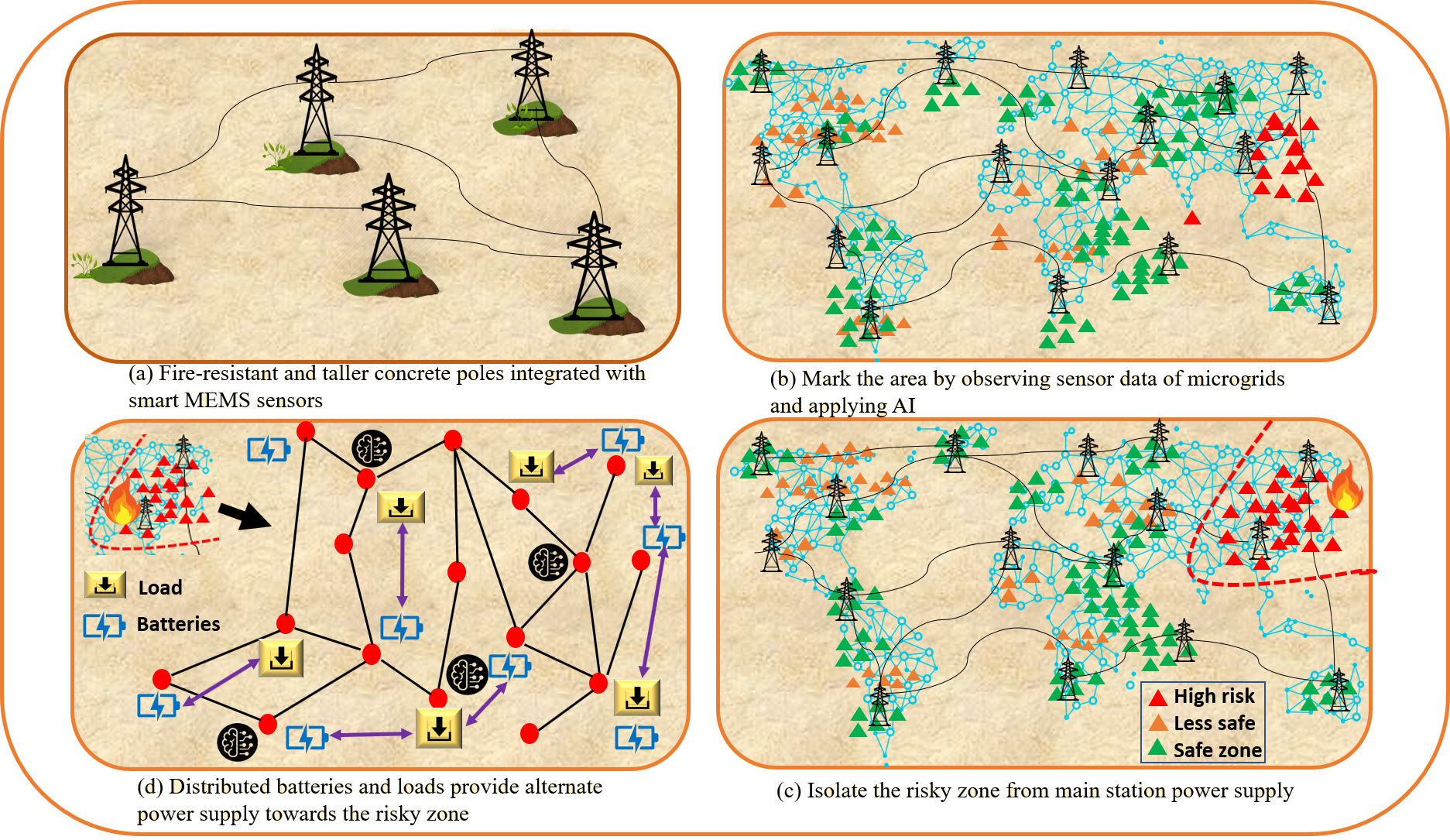}
    \caption{Artificial intelligence for fast recovery after natural hazards, Scenario I: Californian Wildfires} 
    \label{fig:2}  
    \end{center}
\end{figure*}

\begin{figure}[!htb]
\setlength{\belowcaptionskip}{-20pt}
\begin{center}
  \includegraphics[width=0.85\linewidth]{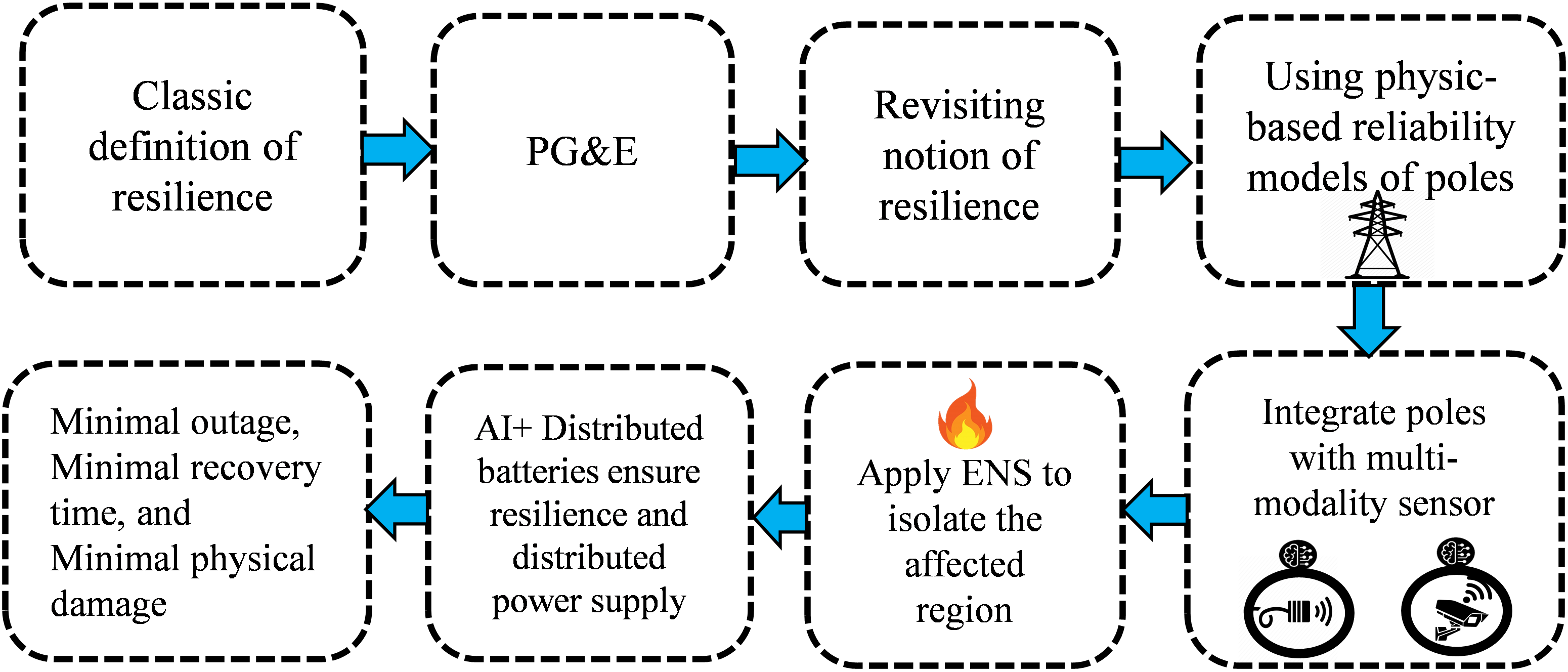}
    \caption{Flowchart of revisiting resilience notion using Artificial Intelligence, Distributed Storage, and Demand Response.}
    \label{fig:3}  
    \end{center}
\end{figure}

\section{classic definition of resilience vs revisited definition based on data revolution}
\subsection{Classic notion of Resilience}
The power grid is such a socio-ecological system that may hold spatial, organization, or temporal parameters that are indirectly affected by society, policy, and economy. Analyzing resilience from different disciplines can help us to build a holistic and all-inclusive definition of resilience. In 1973, Holling \cite{holling} stated that resilience is the ability of a system to maintain its functionality and characteristics after a disturbance. The author in \cite{brown2006defending} inspected resilience from the perspective of infrastructure systems and stated that it is the ability to mitigate the magnitude and duration of disturbances. Resilience can also be seen as the ability to resist stresses and disturbances that are caused by social, economic, and political changes \cite{perrings2007future}. From economic vantage, resilience is the response to economic or environment shocks and facilitate community and people to mitigate losses \cite{adger2000social}. Therefore, the resilience of a system that can be presented by a set of unexpected disturbances is the ability to tackle outage or interruption and recover from perturbation by altering its structure in an agile way. 
\subsection{Resilience Management System's (RMS) risk-aware power distribution scheme}
Our main goal is to maximize the power supply for the RMS and minimize the loss due to the power outages. We formulate 
(1) as a convex optimization problem with control signals and we consider load and battery as our feasibility requirements which can be expressed as, $x_j(t)\hspace{0.1cm} \epsilon \hspace{0.1cm} \chi_{j}(t),\hspace{0.1cm} j\hspace{0.1cm} \epsilon\hspace{0.1cm} B (battery)$ and $x_i(t)\hspace{0.1cm} \epsilon \hspace{0.1cm} \chi_{i}(t),\hspace{0.1cm} i \hspace{0.1cm}\epsilon \hspace{0.1cm}L (load)$.
We want to maximize the RMS utilization which can be considered as single-level convex optimization problem and can be written as follows: 
\begin{equation}
\max_{x(t)} \psi^{RMS} (x(t)) \\
\end{equation}
such that,
\begin{equation}
\sum_{b \epsilon B \cup L} A_{b}x_{b}(t) - c \preceq 0, 
\end{equation}
\begin{equation}
x_j(t)\hspace{0.1cm} \epsilon \hspace{0.1cm} \chi_{j}(t),  \hspace{0.7cm}j \hspace{0.1cm}\epsilon \hspace{0.1cm}B,
\end{equation}
\begin{equation}
x_i(t) \hspace{0.1cm} \epsilon \hspace{0.1cm}\chi_{i}(t), \hspace{0.7cm}i \hspace{0.1cm}\epsilon \hspace{0.1cm}L.
\end{equation}
 
\subsection{RMS's update}
The RMS receives updated vector $z^m(t)$ from the entities including storage and demand agents. Let, $\lambda^m(t)= \overbar{\mathbb{\lambda}}^m(t)$, $\underline{\lambda}^m(t)$, ${\gamma}^m(t)$, ${\beta}^m(t)$ denote the vector of grid-wide dual variables in iteration $m$. The RMS updates $\lambda^m(t)$ is follows: \\
\begin{equation}
\lambda^{m+1}(t)= \Bigg[\lambda^{m}(t) + \zeta^m \sum_{b \epsilon I \cup B}(A_{b}x_b^{m}(t)-c)\Bigg]^{+}
\end{equation}
where, $[.]^+$  is the projection onto the non-negative orthant and $\zeta^a$ is the stepsize in iteration $m$.

\section{Overview of Wildfire in California and Hurricane in Florida, and avoiding interruption while maintaining resilience using AI}
In California, broken out of wildfire is a common phenomenon for the people living there. The most destructive and deadliest wildfire ever recorded in that state happened in the previous year during the 2018 wildfire season. In total, 8,527 fires burnt a major portion of the whole state which is about 1,893,913 acres and it was the largest burning area recorded in a wildfire season \cite{wikiCal}. In that season, Cal Fire spent nearly \$432 million on their operation and \$12 billion insurance was claimed which were mostly due to wildfire destruction \cite{wikiCal}. On August 4, 2018,  a series of large wildfires erupted across California and a national disaster was declared in Northern California, due to the extensive wildfires burning there. In November 2018, another phase of exasperated strong winds took place which caused devastating fires across the state, killed about 85 people and damaged 18,000 structures. According to an analysis that considered U.S. Forest Service data with Zillow housing data stated that nearly half a million homes in California are at risk of wildfire. In California, there are approximately 477,039 homes worth of \$268.2 billion residing in the areas which have “high” and “very high” wildfire hazard potential \cite{Zillow}. Further, Florida is the most storm-affected state in the US due to regular hurricane occurrence and since 1851, there were only 18 hurricane seasons that were passed without having any storm impact. Such storm events resulted in over 10,000 deaths and a huge loss of property \cite{rogers2013noaa}. 

\begin{figure}[!htb]
\setlength{\belowcaptionskip}{-20pt}
\begin{center}
  \includegraphics[width=0.84\linewidth]{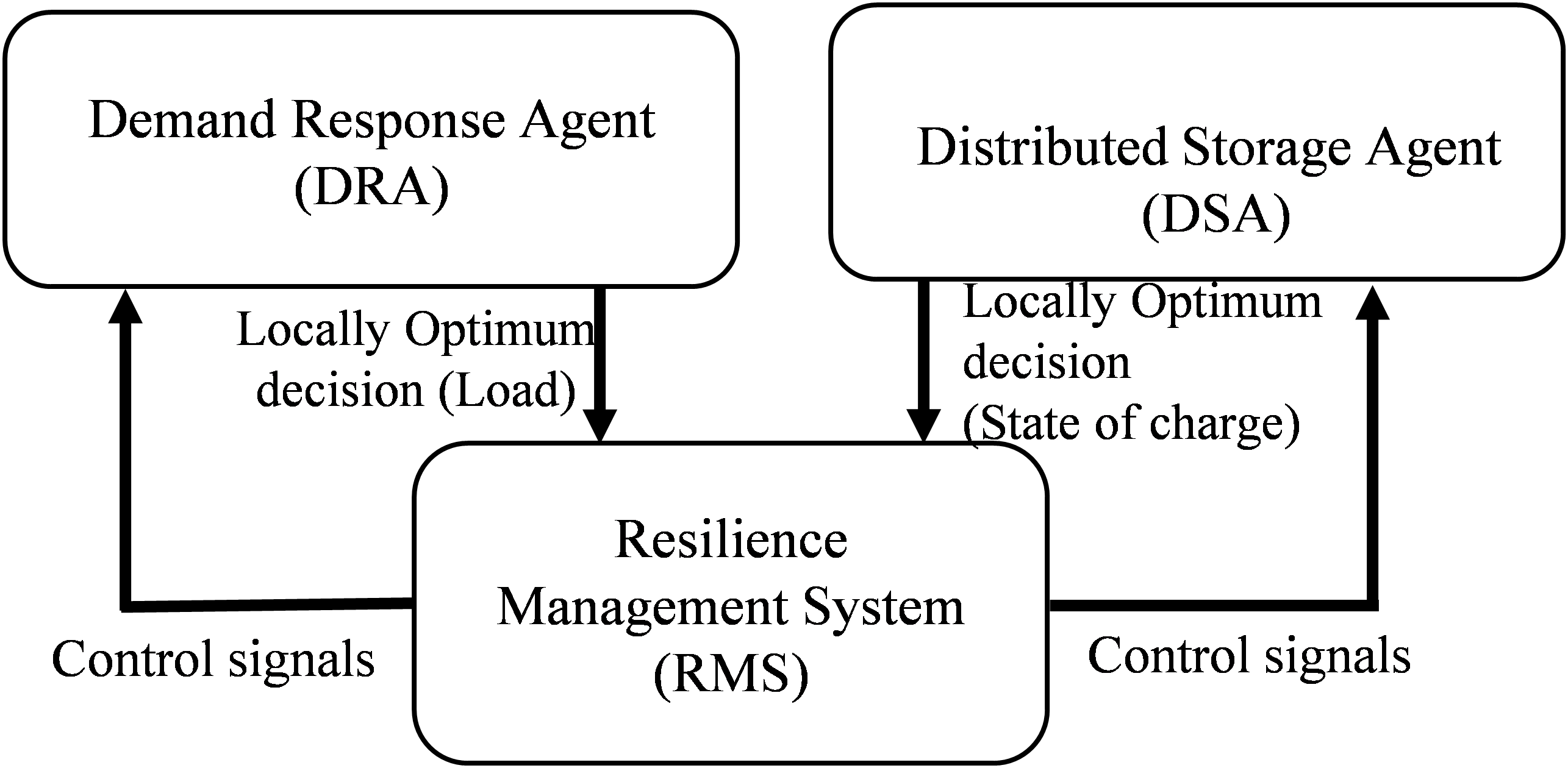}
    \caption{Interactions among demand response agent, distributed storage agent, and Resilience Management System (RMS)}
    \label{fig:4}  
    \end{center}
\end{figure}

The outbreak of such natural disasters is the primary reason for a power outage and besides, some power grid companies are responsible for spreading fire because of their design and placement policy regarding microgrid poles. The fire can be initiated because of falling trees on the poles, damaged or low-quality wires, and sub-optimal deployment of microgrids. We can take more precautions and preventive actions within the regions that are more prone to wildfire (i.e. California) and hurricane (i.e. Florida), hence designing fire placing maps in investigating the route of probable fire spreading regions. This identification of the foreseeable affected region can be done by placing the camera and sensors that can observe and sense the environment and all these components need to be connected with the server. For instance, we assume that all the nodes can communicate with their neighbor nodes and if a component is damaged due to hazards or other incidents, neighboring components can notify the server about the surrounding information. The reading of sensor and camera data can assist us not only to identify the affected region but also the probable regions that are at risk to be affected immediately. The overall system architecture and information flow are depicted in Fig. \ref{fig:2} and Fig. \ref{fig:3} respectively.  \\
After the identification, the regions which are at high risk to be affected imminently can be isolated by stopping power supply from the main station, but, there remains a huge loss due to sudden power outage which needs to be evaluated. To this end, we devise an approach of distributed power supply to handle the power loss due to such catastrophe. We considered two agents in our proposed system model (Fig. \ref{fig:4}) which are demand response agent (DRA) and distributed storage agent (DSA). The purpose of DRA is to track the power load of each region of the affected isolated area and manage the consensus about providing incentives upon receiving power from distributed storage agent. The DSA is such an agent which can communicate with the Resilience Management System (RMS) about its existing power storage and allow RMS to know about its interest in sharing power with other distributed region. The DSA is also responsible for acknowledging the state of charge of the distributed battery, and in case of having enough storage power as well as solidarity in attaining incentives, it gives a positive response to the RMS. The RMS takes care of conveying information among the agents and handles synchronization between the agents. It handles the consensus during the power demand and power supply, and corroborates atomicity i.e. if an agent agrees to share its power resource to other agents, but does not receive desire incentives, then supplied power is not shared with that agent. In this model, local optimum decision will be made in power supply i.e. if there are multiple interested agents requesting power and a specific region demands for 100\% power supply, though 60\% power is enough at that time, then RMS utilizes the rest of the 40\% to other regions, making it optimal for that time convenience. In this way, the resilience of load is achieved through distributed storage agent and thus, the loss due to outage is mitigated by supplying alternate power in a distributed way.

\section{ Leveraging distributed energy storage and edge intelligence to improve reliance}
In a smart grid, the operator is responsible for supplying power energy to all the connected loads. Any interruption or occurrence of an outage in the connected grids can lead to disruption of power supply to the connected nodes. We tailored a network of distributed energy storage that is integrated into all connected nodes concerning the main power grid. Each node has the intelligence to use and check its battery power immediately after getting disconnected from the main station, use alternate battery power in case of a power outage, and recharge its battery from the nearby distributed storage agent during a power outage.  
\\
In algorithm 1, we presented a decentralized AI-based energy market trading solution. When the system just started and in state of its first iteration, we initialized the convergence threshold, the load profile and conventional state of charge profiles in line \textbf{1-4}. In \textbf{5-6}, for time slot $t-1$, we updated the decision variables of DRA, DSA and RMS considering their latest values. In \textbf{8-9}, the connected demand response agent and distributed storage agent dispatch their load and battery charge state to the RMS. Using those, RMS obtains the vector of grid-wide dual variables for iteration $m+1$. In \textbf{10-12}, the RMS sends the control signal to the DRA and DSA with the updated values about the request of power supply quantity and related incentives. In \textbf{11-12}, storage and demand response agents update their decision vector and load profile. From \textbf{13-14}, the storage state of charge is checked whether it is exceeding maximum storage and interested to supply power to the neighbor node. If so, that agent will be appended to the DSA. In \textbf{15-18}, the load of the agent is compared whether it has a resource below a pre-set threshold and if it returns true, then the battery charge request will be sent to RMS. RMS immediately sends an agreement notification by stating the incentives, amount of power to supply and the DSA and DRA profiles to both ends. We repeat this process for each time slot until the difference of voltage phase angle of two consecutive steps reaches a convergence.

{\fontfamily{times}\selectfont
\begin{algorithm}
\DontPrintSemicolon
\caption{\textbf{Distributed Energy Market Trading Approach}}\label{alg:encryption}
Set $m$ := 1 and $\xi$ := $10^{-3}$ \\
\If {t=1}
{
Each demand response agent i $\epsilon$ I randomly initializes its user's appliances load profile \;
Each distributed storage  j $\epsilon$ B randomly initializes its conventional state of charge profiles $p_j^{con,1}(t)$ and $q_j^{con,1}(t)$
}
\uElseIf{t $>$ 1}
{
Demand response agent, distributed storage agent and resilience management system (RMS) initialize their decision variables with their most updated values in the equilibrium at time slot $t-1$
}
\textbf{Repeat} \;
Each demand response agent $i$ and energy storage $j$ send its load profile and state of charge \;
RMS obtains vector $\lambda^{m+1}(t)$ using 
$\lambda^{m+1}(t)= \Bigg[\lambda^{m}(t) + \zeta^m \sum_{b \epsilon I \cup B}(A_{b}x_b^{m}(t)-c)\Bigg]^{+}$

RMS computes the updated values for DSA and DRA, and sends the control signals to the corresponding entity in each bus. \;
Each storage j updates its decision vector. \;
Each demand response agent i updates its load profile. \;
\If{$j$ $>$ $MaxStorage$ and $Interested[j]$ is True}
{Append $Storage_j$ as DSA for $time_t$}
\uElseIf{$i$ $<$ $threshold_{minLoad}$}{Battery charge request to RMS and it immediately responded to DRA for attaining consensus on a $incentives_k$\;
\If{$Interested[i]$ is True on $incentives_k$}{RMS ensures atomicity of transaction and power supply between DRA and DSA}
}
$m : = m+1$. The step size is updated. \;
\textbf{Until} $| x^{m}(t) - x^{m-1}(t) | \leq \xi$

\end{algorithm}
}

\begin{figure}[htb!]
\setlength{\belowcaptionskip}{-20pt}
\begin{center}
  \includegraphics[width=0.82\linewidth]{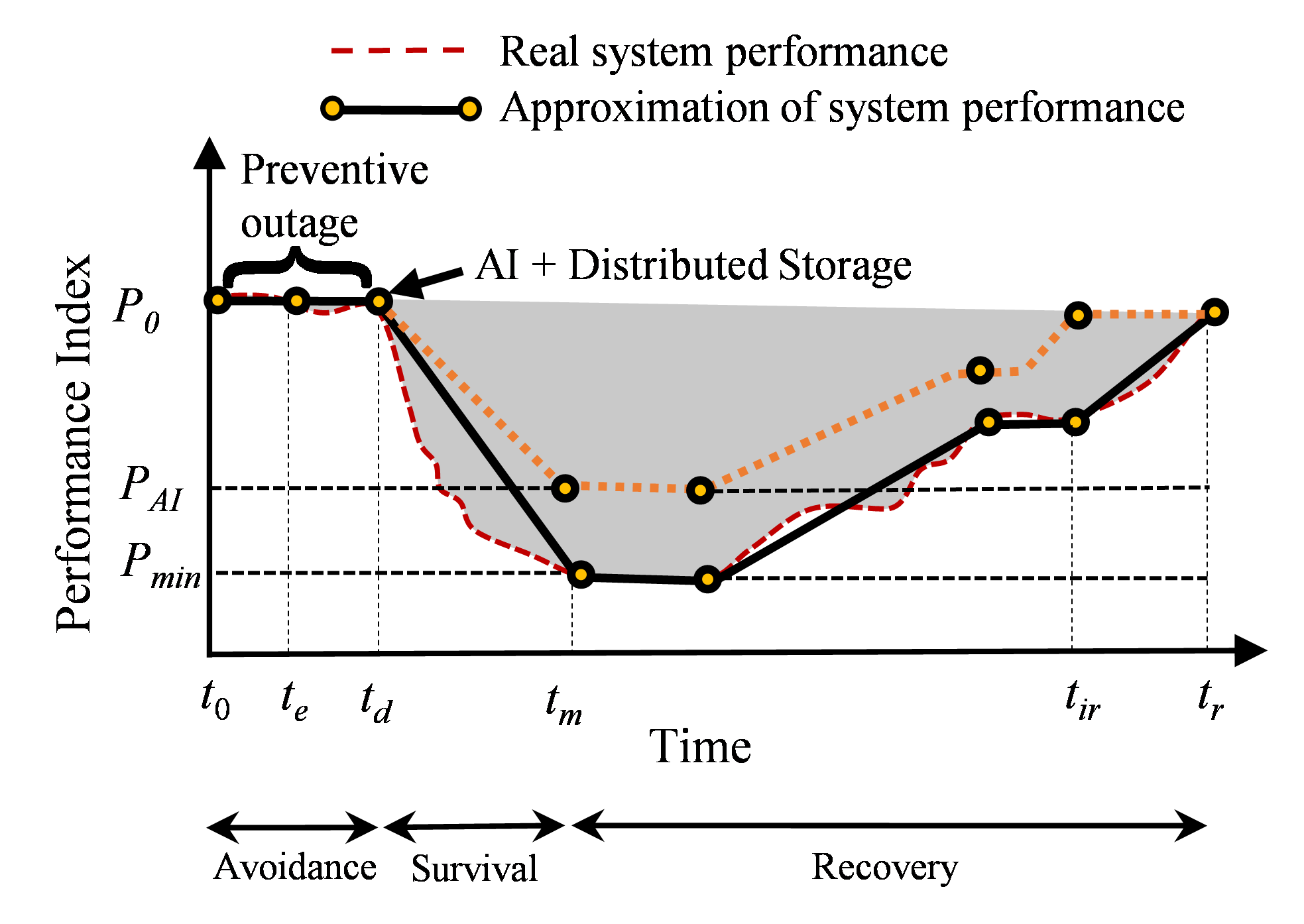}
    \caption{Eight-point linear approximation of system performance in response to an HR event.}
    \label{fig:5}  
    \end{center}
\end{figure}

In Fig. \ref{fig:5}, we presented a linear approximation about system performance to an HR event and can observe that, due to power interruption, the performance index decreased after time $t_d$ and reached at a minimal performance at time $t_m$. Initially, due to preventive outage, the performance index remains at a good point. As soon as the preventive outage stage is finished, the performance index started to going downwards. It reaches a minimum point ($P_{min}$) and started improving its performance index after a certain time interval. On the other hand, when we are integrating the distributed storage with intelligence at the edge nodes, then a sudden power outage does not drag down the performance index to minimum point $P_{min}$; rather than due to resilience, the affected region will be survived by avoiding power outage and significant improvement in performance index can be observed. It can regain its previous performance index by minimizing power loss and enhancing power delivery. This can be obtained using AI for decentralized management of distributed energy storage and demand response.

\section{Conclusion}
In this paper, we articulated the significance of financial burden of wide spread power disruption. This paper also justifies the need for integrating decentralized intelligence towards mitigating consequences of natural disasters. We presented a method for identifying the risk that each geographical region is facing and proposed immediate necessary precaution to reduce this risk. Further, we presented the concept of isolating the affected region by halting power supply from the main station but ensuring the continuation of the power supply using distributed battery and demand response providers. Furthermore, we formulated a decentralized algorithm to establish a consortium among the agents and manage the process of updating the load and battery state of involved entities. Finally, using an eight-point linear approximation, we showed that the integration of artificial intelligence and distributed storage can effectively stabilize performance index in a short amount of time, hence reducing financial loss caused by power outage. Our future work plan includes building on the introduced concept, implementing a decentralized learning-based solution and studying impacts of various natural hazards. 

\bibliographystyle{IEEEtran}
\bibliography{ref}

\end{document}